\documentclass{edm_article}
\begin{document}

\title{Math Multiple Choice Question Generation via Human-Large Language Model Collaboration}

\numberofauthors{4}
\author{
Jaewook Lee$^{1}$,  Digory Smith$^2$, Simon Woodhead$^2$, Andrew Lan$^1$\\
       \affaddr{University of Massachusetts Amherst$^1$}, \affaddr{Eedi$^2$}\\
       \email{\texttt{\{jaewooklee,andrewlan\}@cs.umass.edu}} \\
       \email{\texttt{\{digory.smith,simon.woodhead\}@eedi.co.uk}}
}

\maketitle

\begin{abstract}
Multiple choice questions (MCQs) are a popular method for evaluating students' knowledge due to their efficiency in administration and grading. Crafting high-quality math MCQs is a labor-intensive process that requires educators to formulate precise stems and plausible distractors. Recent advances in large language models (LLMs) have sparked interest in automating MCQ creation, but challenges persist in ensuring mathematical accuracy and addressing student errors. This paper introduces a prototype tool designed to facilitate collaboration between LLMs and educators for streamlining the math MCQ generation process. We conduct a pilot study involving math educators to investigate how the tool can help them simplify the process of crafting high-quality math MCQs. We found that while LLMs can generate well-formulated question stems, their ability to generate distractors that capture common student errors and misconceptions is limited. Nevertheless, a human-AI collaboration has the potential to enhance the efficiency and effectiveness of MCQ generation.
\end{abstract}


\section{Introduction}
Multiple choice questions (MCQs) are widely used to evaluate students' knowledge since they enable quick and accurate administration and grading~\cite{airasian2001classroom,kubiszyn2016educational,nitko1996educational}. MCQs are constructed in a specific format. The \textit{stem} refers to the statement on the problem setup and context, followed by a question that needs to be answered. Among the options, the correct one can be referred to as the \textit{key}, while incorrect ones can be referred to as \textit{distractors}. As the name implies, distractors in MCQs are typically formulated to align with common errors among students. These distractors are chosen because students either i) lack the necessary comprehension of the \textit{knowledge components (KCs)} or concepts/skills tested in the question to accurately identify the key as the correct answer or ii) exhibit misconceptions that make them think a specific distractor is correct.


While MCQs offer many advantages in student knowledge assessment, manually crafting high-quality MCQs, especially in math-related domains, is a demanding and labor-intensive process~\cite{kelly2013adding}. 
There are three main tasks in this process: First, educators need to formulate a question stem that effectively encapsulates the KCs they aim to test. Second, educators need to anticipate common errors and/or misconceptions among students and create corresponding distractors. Third, educators need to provide feedback to students who select distractors that can help them identify their errors and lead them to the correct answer, to expedite their learning process.

The emergence of large language models (LLMs) has raised hopes for making MCQ creation more scalable by automating the process. Specifically, few-shot, in-context learning is promising for generating math MCQs since LLMs can follow instructions based on contextual information conveyed by a few examples. While automated question generation for open-ended questions has shown notable success, generating plausible distractors within MCQs presents a different challenge: distractors should be based on anticipated student errors/misconceptions~\cite{shin2019multiple}, whereas LLMs have not necessarily learned this information during training. Moreover, math MCQs are challenging since they require mathematical reasoning, which means that distractors cannot be generated using a knowledge graph~\cite{stasaski2017multiple} or paraphrasing tool~\cite{le2014automatic}. Consequently, math educators need to take an important role in guiding LLMs in math MCQ generation: LLMs are responsible for scaling up the process while humans use their expertise efficiently. Therefore, we raise following are two core research questions (RQs) that help identify opportunities to generate math MCQs through collaboration between LLMs and human educators: 1) RQ1: Can LLMs generate valid MCQs, especially distractors and feedback corresponding to common student errors/misconceptions? 2) RQ2: What are the key design elements in a system where human math educators and LLMs collaborate on MCQ generation?


\subsection{Contributions}
In this paper, we introduce a prototype tool called the Human Enhanced Distractor Generation Engine(\texttt{HEDGE}) for math MCQ creation, which leverages the expertise of educators by asking them to edit LLM-generated MCQs in a two-step process. In the first step, we prompt the LLM to generate stem, key, and explanation in an MCQ, and ask educators to evaluate and edit the output to make sure it is mathematically correct and relevant to the intended KC. In the second step, we prompt the LLM to generate a set of possible errors/misconceptions and the corresponding distractors and feedback, and ask educators to evaluate and edit the output to make sure they correspond to valid distractors to the generated question stem. In a pilot study, we recruit four former/current math teachers to evaluate our tool on generating math MCQs related to five pre-defined KCs. Results show that educators considered 70\% of the generated stem, key, and explanation generated by GPT-4 as valid. However, they only considered 37\% of the generated misconception, distractor, and feedback valid, which reveals significant limitations of LLMs in capturing anticipated common errors/misconceptions among real students. This observation underscores the necessity of involving humans in the process of generating math MCQs and leveraging real math educators' expertise on common errors among students.

\section{Human Enhanced Distractor Generation Engine} 

\subsection{Overview}
\label{sec:hedge_overview}

\begin{figure}[h]
  \centering
  \includegraphics[width=\linewidth]{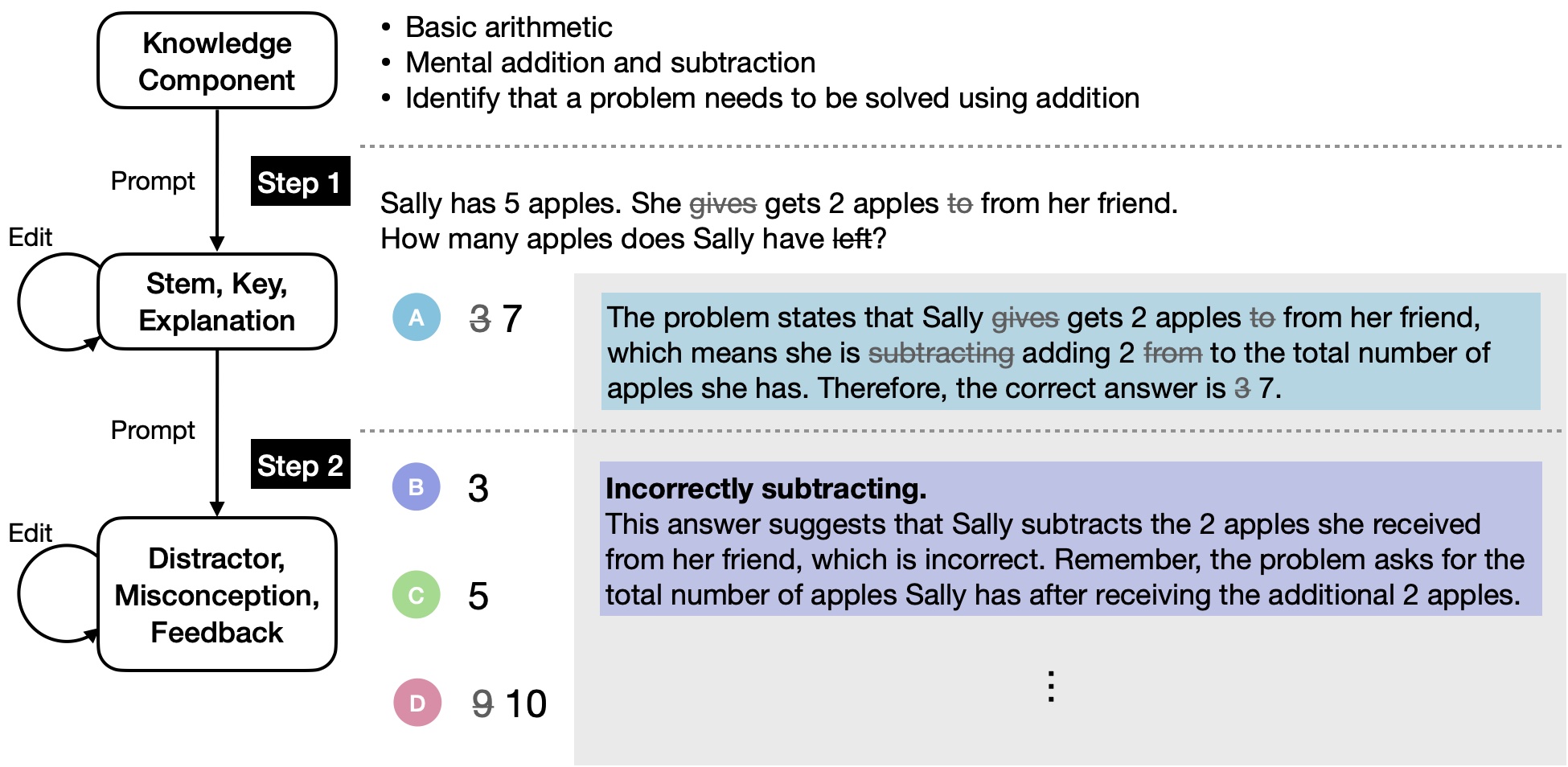}
  \caption{\texttt{HEDGE} Overview: the human-AI collaboration setting for generating math MCQs for a given KC. Strikethrough text represents edits made to LLM-generated content while boldface text indicates misconceptions that correspond to distractors.}
  \label{fig:hedge_overview}
\end{figure}

\texttt{HEDGE} is our prototype for math MCQ generation that generates math MCQ for a given mathematical KC, as illustrated in Figure~\ref{fig:hedge_overview}. These KCs are categorized into three levels of granularity: coarse, medium, and fine-grained. For instance, KCs can cover either a broad topic such as ``basic arithmetic'' or a specific topic like ``Identify that a problem needs to be solved using addition.'' \texttt{HEDGE} is designed to utilize LLMs within OpenAI. The provided example is generated using ChatGPT. We take a two-step approach for MCQ generation: 1) generate the question step and answer key, and an explanation, and 2) generate a list of possible misconceptions, corresponding distractors, and feedback messages. We implement both steps using by prompting LLMs with an in-context example of these tasks.

\begin{table}[h]
\caption{The in-context example used for prompting LLMs for math MCQ generation.}
\centering
\scalebox{.8}{
\begin{tabular}{|c|p{7cm}|}
\hline
\textbf{KC} & Coarse - Ratio, Medium - Writing ratios, Fine - Convert ratios to fractions \\
\hline
\textbf{Stem} & Kate and Isaac share yogurt in a $2:5$ ratio. Kate has $\square$ of the total. Identify the fraction. \\
\hline
\textbf{Key} & $\frac{2}{7}$ \\
\hline
\textbf{Explanation} & The total ratio is 7 parts. Kate's share of $\frac{2}{7}$ is derived by dividing her 2 parts by the total. \\
\hline
\textbf{Misconceptions} & 1. Misinterpreting the ratio as a fraction. \\
& 2. Confusing the difference in ratio parts as relevant. \\
& 3. Calculating Isaac's share instead of Kate's. \\
\hline
\textbf{Distractors} & 1. $\frac{2}{5}$ \hspace{.5cm} 2. $\frac{3}{7}$ \hspace{.5cm} 3. $\frac{5}{7}$ \\
\hline
\textbf{Feedback} & 1. The ratio $2:5$ means 7 parts total, not $\frac{2}{5}$. \\
& 2. The ratio splits the total, not the difference between parts. \\
& 3. Ensure you are calculating Kate's share, not Isaac's. \\
\hline
\end{tabular}
}
\label{tab:in_context_example}
\end{table}

The in-context example shows the KC converting ratios to fractions, employing a real-life scenario in which Kate and Isaac share yogurt in a $2:5$ ratio. The objective is to calculate the fraction representing Kate's share, $\frac{2}{7}$. In this context, we list three common misconceptions. First, a student mistakenly thinks that the ratio $2:5$ could be directly converted into the fraction $\frac{2}{5}$. Second, a student mistakenly calculates the difference between Kate's and Issac's share. Third, a student mistakenly think the goal is to calculate Issac's share. These misconceptions, along with the corresponding feedback on how to resolve them, are included as part of the in-context example. 

Now, we explore a scenario where an educator creates MCQs using our tool based on the concept of basic arithmetic, specifically focusing on mental addition. In the first step, given the target KC, along with an in-context example consisting of the concept, stem, key, and explanation, the LLM generates the following stem: ``Sally has 5 apples. She gives 2 apples to her friend. How many apples does Sally have left?'' However, this stem mistakenly embodies the KC of subtraction rather than addition. Therefore, the educator edits the generated results to align it with the intended KC of addition. In the second step, using the adjusted stem, key, and explanation, as well as incorporating in-context examples with distractors, misconceptions, and feedback, the LLM generates distractors along with corresponding misconceptions and feedback. Figure~\ref{fig:hedge_overview} illustrates option $B$, which contains a misconception related to subtraction instead of addition, accompanied by feedback designed to correct this error. Additionally, the educator has the option to edit option $D$ to address any misconceptions associated with multiplication.

\subsection{User Interface}

\begin{figure}[h]
  \centering
  \includegraphics[width=\linewidth]{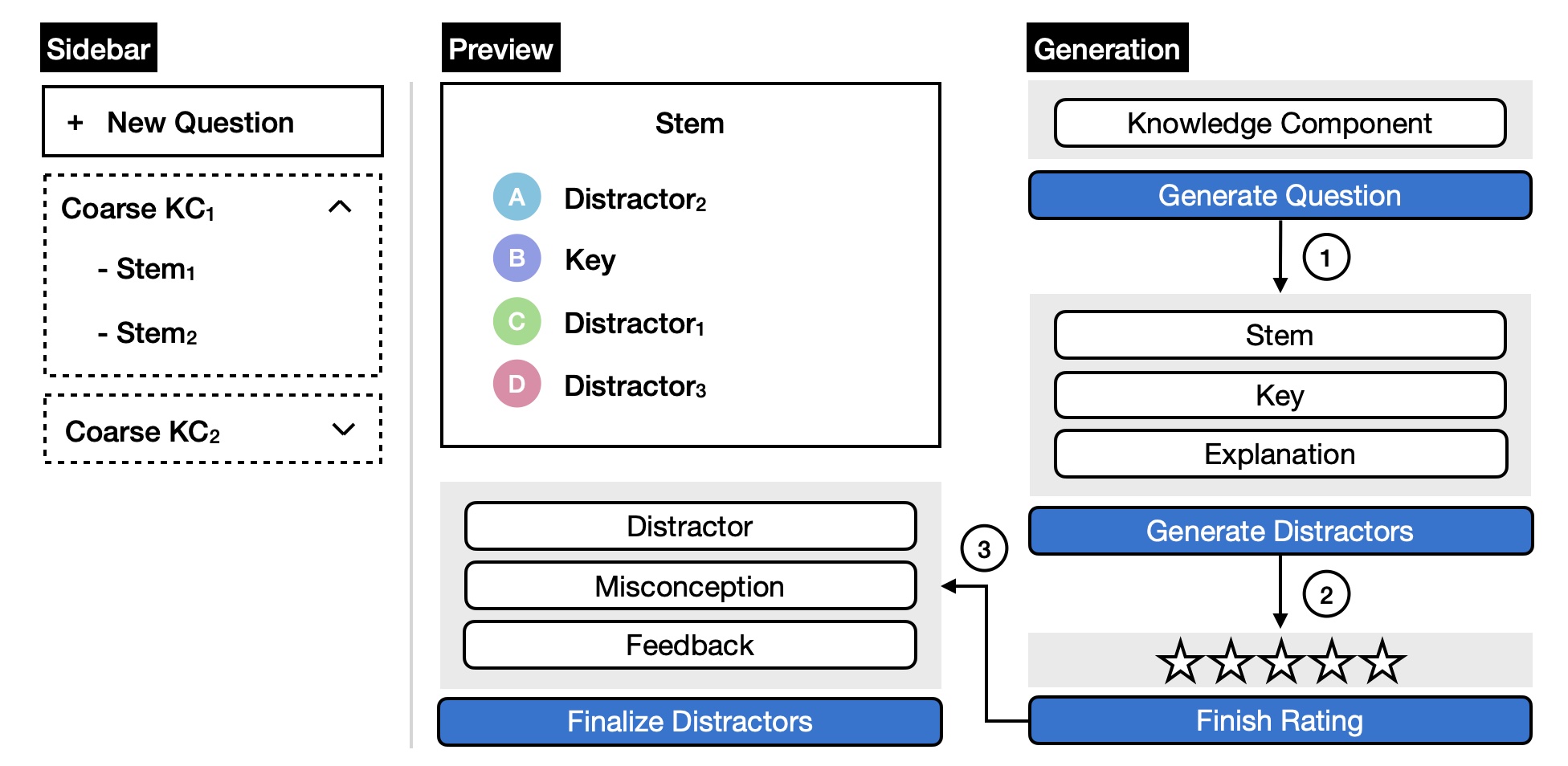}
  \caption{\texttt{HEDGE} Interface: what human participants use to generating an MCQ by editing LLM output.}
  \label{fig:hedge_interface}
\end{figure}

We develop \texttt{HEDGE} interface, as illustrated in Figure~\ref{fig:hedge_interface}. This interface is built using React and employs Firestore as its database for data storage. The interface comprises three components: a Sidebar, a Preview, and a Generation.

The educator generates MCQs using the Generation component as discussed in Section~\ref{sec:hedge_overview}. Here, after prompting LLMs using the edited stem, key, and explanation, we add a rating step to assess the overall quality of misconceptions, distractors, and feedback that the educator rates based on a 5-point Likert scale. 

Once the educator completes the distractor editing process, the Preview component displays a fully structured MCQ, with the answer options randomized. We store any metadata that isn't visually represented within the image. Following the completion of distractor editing, the Sidebar component is refreshed. The educator can click on the stem to view the generated image along with the answer sheet or create a new MCQ.

\section{Pilot Study}


\begin{table*}
\caption{Pre-defined math KCs used in the pilot study.}
\centering
\scalebox{.75}{
\begin{tabular}{|l|l|l|}
\hline
\textbf{Coarse-grained}                        & \textbf{Medium-grained}                                     & \textbf{Fine-grained}                                                                    \\ \hline
Factors, Multiples and Primes & Factors and Highest Common Factor          & Identify factors of a number                                            \\ \hline
Fractions                     & Equivalent Fractions                       & Identify equivalent fractions when presented as numbers                 \\ \hline
Indices, Powers and Roots     & Squares, Cubes, etc                        & Understand the notation for powers                                      \\ \hline
Percentages                   & Repeated Percentages and Compound Interest & Understand the elements of the formula for compound percentage decrease \\ \hline
Surds                         & Simplifying Surds                          & Write a simplified surd in a non-simplified form                        \\ \hline
\end{tabular}
}
\label{tab:construct}
\end{table*}

\subsection{Experimental Setup}
We perform a pilot study to assess the usability of \texttt{HEDGE} in generating MCQs. In this study, we select pre-defined KCs and instruct participants to utilize these KCs to simulate a scenario where an educator is crafting MCQs. We select the KCs and the in-context example from a large education company's content repository, categorized under the label ``Number,'' encompasses various subtopics, such as ``Basic Arithmetic,'' ``Fractions,'' and ``Rounding and Estimating.'' We choose five KCs, as shown in Table~\ref{tab:construct}, from the KCs that incorporate mathematical expressions, such as fractions, powers, and surds. We utilize GPT-4 as LLM for the study and set the parameters to temperature $=0.7$ and top\_p value $=0.9$ to balance creativity and consistency of the generated MCQs. After completing the study, participants are asked to complete an exit survey. The survey includes open-ended questions and ratings on their satisfaction with the quality of LLM-generated responses and the usability of the tool using a 5-point Likert scale.

\subsection{Participants}
We recruit four participants for the study, comprising one male and three females, all recruited through Upwork~\cite{upwork}. Among them, two currently work as middle/high school math teachers, while the other two currently work as tutors, with prior experience as former math teachers. All participants are selected based on their qualifications and expertise in mathematics education. Each participant was tasked with creating five MCQs using the \texttt{HEDGE}, employing the five KCs specified in Table~\ref{tab:construct}.

\begin{table}[ht!]
\caption{Question stems generated using \texttt{HEDGE} and the corresponding KCs.}
\centering
\scalebox{.75}{
\begin{tabular}{|p{4cm}|p{6cm}|}
\hline
\textbf{Fine-grained KC}  & \textbf{Stem} \\ \hline
$1$. Identify factors of a number \newline \textit{Which of these numbers is not a factor of 9?}   & $a$. What are all the factors of the number 12? \newline $b$. What are the factors of $18$? \newline $c$. Which of the following is a factor of 18? \newline $d$. Which of the following numbers is a factor of 36? \\ \hline
$2$. Identify equivalent fractions when presented as numbers \newline \textit{ Which fraction is equivalent to $\frac{9}{13}$?} & $a$. Sue has a fraction of $\frac{4}{8}$. What fraction is equivalent to the fraction she has? \newline $b$. The fraction $\frac{6}{18}$ is equivalent to which of the following fractions? \newline $c$. Which of the following fractions is equivalent to $\frac{3}{9}$? \newline $d$. Which of the following fractions is equivalent to $\frac{2}{4}$? \\ \hline
$3$. Understand the notation for powers \newline \textit{ To calculate ${53}^2$ you need to do ...} & $a$. The number $3^2$ is equal to $\square$. What number completes the sentence? \newline $b$. The number $3^4$ represents $\square$. What number completes the sentence? \newline $c$. If $a^3$ is read as "a cubed", how is $a^4$ read? \newline $d$. What is the value of $2^{3}$? \\ \hline
$4$. Understand the elements of the formula for compound percentage decrease \newline \textit{ A car depreciates in value by 10\% each year. If a car was bought for \$4500, what calculation would find the value of the car after 3 years?} & $a$. A car that costs \$5000 loses 12\% of its value each year. After one year, the car is worth $\square$. What completes the sentence? \newline $b$. A new car loses 20\% of its value each year. If the car was originally priced at \$15,000, what will be its value after 2 years? \newline $c$. The price of a car is reduced by 5\%  each year. If the car was originally priced at \$5000, what will be the price of the car after two years? \newline $d$. A car depreciates in value by 10\% each year. If the car is initially worth \$35000, what is the formula to calculate the car's value after n years? \\ \hline
$5$. Write a simplified surd in a non-simplified form \newline \textit{$5\sqrt{13} = \sqrt{n}$ What is the value of $n$?} & $a$. If $2\sqrt{5}$ is a simplified surd, what is its non-simplified form? \newline $b$. The square root of 18 is written in simplified surd form as $3\sqrt{2}$. How can it be rewritten in a non-simplified form? \newline $c$. Simplify the surd $\sqrt{45}$. \newline $d$. A non-simplified surd is $\sqrt{8}$. How can it be represented in simplified form? \\ \hline
\end{tabular}
}
\label{tab:stem_result}
\end{table}

\section{Results}

\subsection{Stem, Key, and Explanation}
\label{sec:result_1}
Table~\ref{tab:stem_result} shows the stems produced by participants utilizing \texttt{HEDGE}. In the ``Fine-grained KC'' column, the original stem is indicated in \textit{italics}, while the stems modified by each participant denoted as $a$, $b$, $c$, and $d$, respectively. In what follows, we label each MCQ in the format of $1a$, where $1$ denotes the index of the fine-grained KC and $a$ denotes index of the participant. 

Out of 20 sets of stem, key, and explanation generated by the LLM, participants deemed 14 sets of them as valid. Among these valid sets, two added more details in their explanations, while the remaining sets were adopted without any need for edits. For example, italicized details were added in the explanation for $2c$: ``The fraction $\frac{3}{9}$ simplifies to $\frac{1}{3}$ because both the numerator and the denominator can be divided by \textit{a common factor of} 3. \textit{3 divided by 3 is 1, and 9 divided by 3 is 3.} Hence, $\frac{1}{3}$ is an equivalent fraction to $\frac{3}{9}$.'' The other case was to make the question setting more realistic: In $4d$, the educator edited the initial price of the car worth \$5000 to \$35000. This adjustment reveals the limitations of LLMs in accurately representing real-life problem scenario. We now analyze the cases that participants deemed invalid. 

\paragraph{Grammar error}
In $2a$, educator corrected grammar error of ``she have'' to ``she has.'' No other grammar errors occurred in the study besides this one, underscoring the capability of LLMs to consistently produce grammatically correct sentences.

\paragraph{Not mastering KC}
Regarding 5th KC, GPT-4 shows a lack of knowledge on the distinction between simplified and non-simplified surd. The followings are invalid stems generated by GPT-4: 1) $5a$. If $\sqrt{20}$ is a simplified surd, what is its non-simplified form? 2) $5c$. Express the simplified surd $\sqrt{45}$ in a non-simplified form. 3) $5d$. A simplified surd is $\sqrt{8}$. How can it be represented in non-simplified form? 

This invalid stem has misled a participant to edit a stem to convey KC as simplifying surd, which is the opposite of non-simplifying surd ($5c$). 

\paragraph{Calculation error}
In $4c$, GPT-4 generated a key of \$4750, erroneously calculating the car price after one year instead of two years. However, in the other three cases within the same KC, GPT-4 calculated correctly, showing its math problem-solving skills. 

\subsection{Distractor, Misconception, and Feedback}
Table~\ref{tab:distractor_result} shows a breakdown of 60 distractors (comprising three distractors for 20 stems), categorized based on the validity of misconceptions, distractors, and feedback. 

\begin{table}
\caption{Breakdown of the 60 generated distractors and their quality ratings. (\cmark: valid, \xmark: invalid)}
\centering
\scalebox{0.8}{
\begin{tabular}{|c|c|c|c|c|c|}
\hline
\textbf{Case} & \textbf{Misconception} & \textbf{Distractor} & \textbf{Feedback} & \textbf{Ratio} & \textbf{Rating}\\ \hline
1 & \cmark & \cmark & \cmark & 37\%  & 4.8\\ \cline{1-1} \cline{4-6} 
2 & & & \xmark & 8\% & 2.8 \\ \cline{1-1} \cline{3-6} 
3 & & \xmark & \cmark & - & -  \\ \cline{1-1} \cline{4-6} 
4 & & & \xmark & 18\% & 2.1 \\ \hline
5 & \xmark & \cmark & \cmark & 12\% & 3.4  \\ \cline{1-1} \cline{4-6} 
6 & & & \xmark & 5\% & 3.0 \\ \cline{1-1} \cline{3-6} 
7 & & \xmark & \cmark & - & -\\ \cline{1-1} \cline{4-6} 
8 & & & \xmark & 20\% & 2.3 \\ \hline
\end{tabular}
}
\label{tab:distractor_result}
\end{table}

\paragraph{Adopt All Responses (Case 1, 37\%)}
Among 60 distractors, educators identified 22 responses as valid, including two cases that are actually invalid.

\paragraph{Edit Feedback Only (Case 2, 8\%)}
These cases have valid misconception and distractor and educators has made adjustments to the feedback to enhance its clarity. For example, one of the distractors for $2d$ is $\frac{2}{3}$. The feedback generated by GPT-4 is as follows: ``You seem to have compared only the numerators of the fractions. However, when checking for equivalent fractions, both the numerator and denominator need to be considered. The fraction $\frac{2}{3}$ is not equivalent to $\frac{2}{4}$.'' The educator removed the redundant final sentence and introduced ``Remember, equivalent fractions require both the numerator and denominator to be proportional.'', which helps students better understand the importance of considering both the numerator and denominator when comparing fractions for equivalence. This adjustment emphasizes that the equivalence between fractions relies on maintaining proportionality between the numerator and denominator. While GPT-4 provides valid explanations, it sometimes fail to include critical insights that are necessary for students' improvement.

\paragraph{Adopt Misconception Only (Case 4, 18\%)}
These cases are often due to a mismatch between the misconception and the distractor. In $4c$, the misconception ``The student mistakenly believed that the car depreciates by a constant amount each year, not a percentage.'' did not match the distractor $35000 - 0.10n$. Additionally, there are cases when, even if the distractor is valid, it may not effectively encapsulate student misconceptions. In $1a$, the educator updated the distractor from $1, 2, 3, 4, 6, 12, 24$ to $12, 24, 36, 48, 60$, making it a more attractive distractor for those who confuse factors for multiples.

\paragraph{Edit Misconception Only (Case 5, 12\%)}
As in Case 4, invalid cases are often due to a mismatch between the misconception and the distractor. In $5d$, the misconception ``The student may believe that all square roots are in their simplest form.'' did not match the distractor ``$\sqrt{2}$.'' The educator updated the misconception as ``The student may have confused square roots with cube roots.'' providing a more accurate misconception for the distractor. Additionally, there are cases when, even if the misconception is valid, it may not likely be the misconception why the student selects the distractor. In $1c$, the educator updated the misconception of distractor ``$4$'' from ``The student might think that only the numbers less than 18 can be the factors of 18.'' to ``The student might think that any even number can be a factor of an even number.'', making it more accurate for addressing the student's misconception.

\paragraph{Adopt Distractor Only (Case 6, 5\%)}
These cases were when educators adopted distractors and edited wrong misconceptions and feedback. For example, in the case of $5a$, $\sqrt{10}$ is a valid distractor as the student could simply multiply $2$ and $5$. However, the misconception and feedback generated by GPT-4 did not align with the distractor; therefore the educator had to edit it accordingly. 

In Cases 4, 5, and 6, LLMs revealed inconsistent mathematical reasoning when analyzing misconceptions, distractors, and feedback for a given stem. The inconsistency underscores a necessity for human educators to manually align distractors and their underlying misconceptions and corresponding feedback in many cases.

\paragraph{Reject All Responses (Case 8, 20\%)}
These cases were when misconceptions had poor quality or were wrong, resulting in inadequate distractors and feedback. Two of the distractors generated for $2b$ by GPT-4 shows both poor quality and wrong misconceptions. While the misconception in the first distractor is valid, stating that ``The student may not divide both the numerator and denominator by the same number,'' the distractor itself, represented by $\frac{3}{9}$, and its associated feedback lack coherence and fail to align with this misconception. Meanwhile, the misconception in the second distractor ($\frac{8}{24}$) lacks coherence, as expressed in the following manner: ``The student may confuse the concept of equivalent fractions with simplifying fractions.'' These results reveal that LLMs often fail to anticipate valid misconceptions and errors that are common among students, making human educators' involvement crucial in the creation of math MCQs. 

\subsection{Takeaways from the Survey}
After the study, participants were asked to fill out the survey, asking the experience using \texttt{HEDGE}. We categorize result into two: Quality of LLM-generated responses and Tool Usability. 

\subsubsection{Quality of LLM-generated responses.}

\paragraph{Stem, Key, and Explanation}
On a 5-point Likert scale, the participants gave an average rating of 4. This rating aligns with the open-ended responses regarding most of the generated stem, key, and explanation valid. However, two participants addressed the tool's limitation in terms of the level of question difficulty. One participant points out that the questions appear to be at a low Bloom's Taxonomy level. For example, ``If $a^3$ is read as `a cubed', how is $a^4$ read?''  While it's important for students to grasp the verbal representation of these terms, educators often place greater emphasis on whether students understand the equivalent expressions and concepts associated with them. The other participant points out that the Depth of Knowledge (DOK) levels predominantly focused on Level 1 (Recall) and Level 2 (Skill or Concept). We can prompt LLMs to generate questions at various Bloom's or DOK levels to enhance the question difficulty and promote deeper understanding~\cite{elkins2023useful}. Moreover, we can invite educators to craft in-context examples with higher Bloom's or DOK levels.

\paragraph{Distractor, Misconception, and Feedback}
On a 5-point Likert scale, the participants gave an average rating of 2.5. This rating aligns with the open-ended responses regarding most of the generated misconceptions, distractors, and feedback that do not reflect what students typically make in the classroom based on the participant's teaching experience. The responses again point to the observation that LLMs do not understand errors that student are likely to make. One participant suggest providing a ``bank'' of misconceptions that educators could refer to. We can prompt LLMs to generate multiple misconceptions and engage educators in ranking these misconceptions based on their alignment with actual student errors.

\subsubsection{Tool Usability}
\paragraph{User Interface}
On a 5-point Likert scale, the participants gave an average rating of 4 for comfort level with generating MCQs using \texttt{HEDGE} while giving an average rating of 3.25 for the effectiveness of generating high-quality MCQs. Participants are enthusiastic about the tool's potential for simplifying the process of generating MCQs but are nevertheless skeptical about LLMs' capability to generate valid distractors. We will need to enhance the tool by making improvements in the quality of generated distractors to align more closely with educators' expectations.

\section{Conclusions and Future Work}
In this paper, we conducted a pilot study using a prototype tool \texttt{HEDGE} to explore the opportunity for collaboration between LLMs and humans in generating math MCQs. We identified that while LLMs can generate valid stems, keys, and explanations, they are currently limited in capturing anticipated student errors, which is reflected in invalid misconceptions, distractors, and feedback.

This study opens up many avenues for future work. First, we can extend the prompt with more in-context examples. Currently, we use only one in-context example; using multiple in-context examples can help guide LLMs to help capture valid misconceptions for the target stem. As mentioned in our survey takeaways, we can also add in-context examples with different Bloom's taxonomy and difficulty levels to enhance the diversity of the generated questions. 
We could also use techniques for optimally selecting these in-context examples~\cite{scarlatos2023reticl}.
Second, we can change the interface to choose generated distractors from a bank that contains more than three distractors. 
When building the bank, we can employ a $k$-nearest neighbor approach that gauges question similarity and leverage LLMs to generate distractors~\cite{feng2024exploring}.
Educators will have less burden on thinking of anticipated misconceptions and even benefit from discovering misconceptions they might have overlooked. Third, improving this tool as a platform so that educators can share their misconceptions will result in a constantly-expanding error bank, which will benefit future MCQ generation. 
Fourth, we can provide educators choice to create personalized questions by adding a module to customize the named entity or topic (e.g., sports, popular culture) in an MCQ to stimulate student interest and make the question more culturally relevant~\cite{walkington2019personalizing}. 
Lastly, we can extend the area of Human-LLM collaboration to other domains beyond math MCQ generation, such as feedback generation~\cite{scarlatos2024improving}, question generation~\cite{kumar2023improving}, and programming education~\cite{ahmed2020synthesizing}.

\section{Acknowledgements}
We thank Schmidt Futures and the NSF (under grants IIS-2118706 and IIS-2237676) for partially supporting this work. 

\clearpage
\bibliographystyle{abbrv}
\bibliography{sigproc}  
\end{document}